%% file: main.tex
\documentclass[10pt,twocolumn,letterpaper]{article}

\usepackage{cvpr}              

\input{preamble}

\definecolor{cvprblue}{rgb}{0.21,0.49,0.74}
\usepackage[pagebackref,breaklinks,colorlinks,citecolor=cvprblue]{hyperref}
\usepackage{colortbl}
\usepackage{moresize}
\def\utc{\stackrel{\rm c}{=}}
\def\uT{u_{\text{\sc t}}}

\title{Approximate Size Targets \\ Are Sufficient for Accurate Semantic Segmentation}

\author{Xingye Fan\\
University of Waterloo\\
{\tt\small x44fan@uwaterloo.ca}
\and
Zhongwen (Rex) Zhang\\
University of Waterloo\\
{\tt\small z889zhan@uwaterloo.ca}
\and
Yuri Boykov\\
University of Waterloo\\
{\tt\small yboykov@uwaterloo.ca}
}

\begin{document}
\maketitle
\input{sec/0_abstract}    
\input{sec/1_intro}

\input{sec/2_properties}
\input{sec/3_experiments}

\input{sec/4_conclusions}
{
    \small
    \bibliographystyle{ieeenat_fullname}
    \bibliography{main}
}

\end{document}

%% file: preamble.tex
\usepackage[dvipsnames]{xcolor}

\newcommand{\confName}{Conference Name} 


%% file: sec/0_abstract.tex
\begin{abstract}
This paper demonstrates a surprising result for segmentation with image-level targets: extending binary class tags to approximate relative object-size distributions allows off-the-shelf architectures to solve the segmentation problem.
A straightforward zero-avoiding KL-divergence loss
for average predictions produces segmentation accuracy comparable to the standard pixel-precise supervision with full ground truth masks. In contrast, current results based on class tags typically require complex non-reproducible architectural modifications and specialized multi-stage training procedures.
Our ideas are validated on PASCAL VOC using our new human annotations of approximate object sizes. We also show the results on COCO and medical data using synthetically corrupted size targets. All standard networks demonstrate robustness to the size targets' errors. For some classes, the validation accuracy is significantly better than the pixel-level supervision; the latter is not robust to errors in the masks. Our work provides new ideas and insights on image-level supervision in segmentation and may encourage other simple general solutions to the problem.
\end{abstract}

%% file: sec/1_intro.tex
\section{Introduction}
\label{sec:introduction}

Our image-level supervision approach to semantic segmentation can be easily
summarized using only a few standard notions. Soft-max predictions $S_p = (S_p^1,\dots,S_p^K)$ 
generated by the segmentation model at any pixel $p$ are categorical distributions over given $K$ classes of objects (including background) present in the given set of training images. At any image, the average prediction is defined as
\begin{equation} \label{eq:ap}
    \bar{S} = \frac{1}{|\Omega|}\sum_{p\in\Omega} S_p 
\end{equation}
where $\Omega$ is a set of pixels $p$ in the given image. The average prediction $\bar{S} =(\bar{S}^1,\dots,\bar{S}^K)$ is 
also a categorical distribution over $K$ classes. It can be seen as an image-level prediction of the relative or normalized sizes (volume, area, or cardinality) 
of the image objects. 
\begin{figure}[!t]
\includegraphics[width=0.98\linewidth]{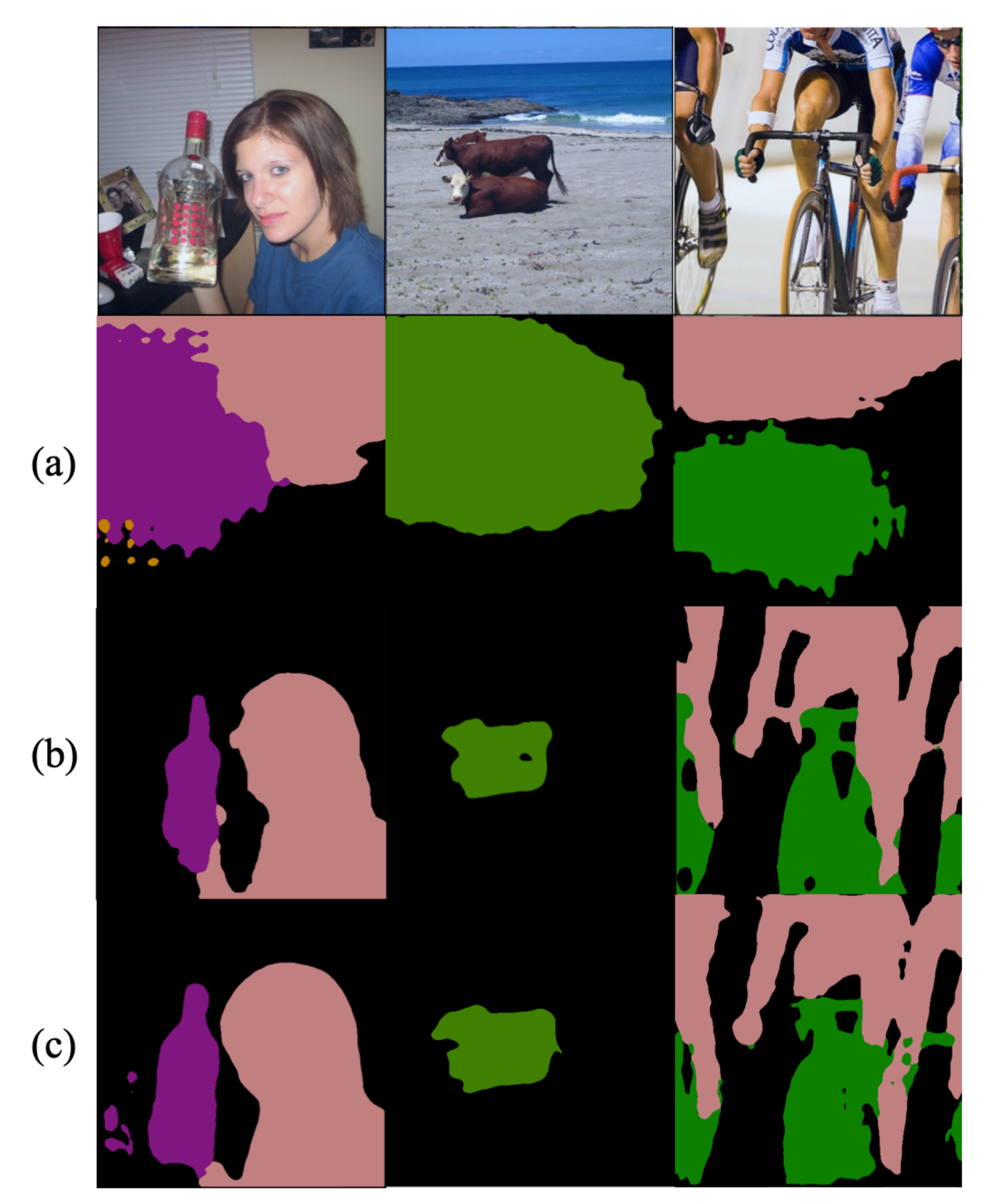} \vspace{-1ex}
    \caption{Semantic segmentation from image-level supervision: test results for training by
    (a) log-barriers \eqref{eq:kolesnikov_expand_loss}  and (b) our approximate size targets \eqref{eq:size_loss}. 
    Full GT-mask supervision results are in (c). \vspace{-3ex} \label{fig:teaser}}  
\end{figure}

We assume training images have approximate size targets represented by categorical distributions $v=(v_k)_{k=1}^K$. For each such image,
our {\em size-target loss} is defined as
\begin{equation} \label{eq:size_loss}
    L_{size} \;\;=\;\;KL(v\|\bar{S})\;\;=\;\;\sum_k  v_k\ln\frac{v_k}{\bar{S}^k}
\end{equation}
where $KL$ is Kullback–Leibler divergence. Figure~\ref{fig:teaser}(b) shows some test results for
a generic segmentation network (WR38 backbone) trained on PASCAL VOC using only image-level supervision 
with approximate size targets ($8\%$ mean relative errors). 
Our total loss is very simple: it combines size-target loss \eqref{eq:size_loss} and a common CRF loss \eqref{eq:CRF_loss}.

\subsection{Overview of weakly-supervised segmentation}
By {\em weakly-supervised} semantic segmentation we refer to all methods that do not use full pixel-precise ground truth
masks for training. Such full supervision is overwhelmingly expensive for segmentation and is
unrealistic for many practical purposes. There are many forms of weak supervision for semantic segmentation, 
e.g. using partial pixel-level ground truth defined by ``seeds'' \cite{lin2016scribblesup,tang2018regularized} or/and image-level supervision by class-tags \cite{papandreou2015weakly,kolesnikov2016seed,araslanov2020single}. It is also common to incorporate self-supervision based on various augmentation ideas and contrastive losses \cite{ji2019invariant,cho2021picie,zhou2022regional}.

Lack of supervision also motivates unsupervised loss functions such as standard old-school 
regularization objectives for {\em low-level} segmentation or clustering. For example, many methods \cite{caron2018deep,hwang2019segsort,zhou2022regional}
use variants of K-means objective (squared errors) enforcing the compactness of each class representation.
It is also very common to use CRF-based pairwise loss functions \cite{lin2016scribblesup,tang2018regularized} 
that encourage segment shape regularity and
alignment to intensity contrast edges in each image \cite{boykov2001interactive}. 
The last point addresses the well-known limitation of
standard segmentation networks that often output low-resolution segments. 
Intensity contrast edges on the high-resolution input image is a good low-level cue of an object boundary and it can improve the details and localization of the semantic segments. 

Conditional or Markov random fields (CRF or MRF) are common basic examples of pairwise graphical models. The corresponding unsupervised loss functions can be formulated for continuous soft-max predictions $S_p$ produced by segmentation networks, e.g. \cite{kolesnikov2016seed,lin2016scribblesup,tang2018regularized}. Thus, it is natural to use relaxations of the standard discrete CRF/MRF models, such as {\em Potts} \cite{boykov2001fast} 
or its {\em dense-CRF} version \cite{krahenbuhl2011efficient}. 
We use a bilinear relaxation of the general Potts model
\begin{equation} \label{eq:CRF_loss}
   L_{crf}(S) \;\;=\;\; \sum_{k} (\mathbf{1}-S^k)^\top W S^k 
\end{equation}
where $S:=(S_p\,|\,p\in\Omega)$ is a field of all pixel-level soft-max predictions $S_p$ in a given image, and $S^k:=(S_p^k\,|\,p\in\Omega)$ is a vector of all pixel predictions specifically 
for class $k$. Matrix $W=[w_{pq}]$ typically represents some given non-negative affinities $w_{pq}$ 
between pairs of pixels $p,q\in \Omega$. It is easy to interpret loss \eqref{eq:CRF_loss} assuming,
for simplicity, that all pixels have confident {\em one-hot} predictions $S_p$ so that each $S^k$ 
is a binary indicator vector for segment $k$. Then the loss sums all weights $w_{pq}$ between the
pixels in different segments. Thus, the weights are interpreted as discontinuity penalties.
The loss minimizes the discontinuity costs \cite{boykov2001fast}.

In practice, affinity weights $w_{pq}$ are set close to $1$ if two neighboring pixels $p,q$ 
have similar intensities, and weight $w_{pq}$ is set close to zero either when two pixels are 
far from each other on the pixel grid or if they have largely different intensities \cite{boykov2001fast,krahenbuhl2011efficient,tang2018regularized}.
The affinity matrix $W$ could be arbitrarily dense or sparse, e.g. many zeros when representing 
a 4-connected pixel grid. The non-zero discontinuity costs between neighboring pixels are
often set by a Gaussian kernel $w_{pq}=\exp\frac{-\|I_p-I_q\|^2}{2\sigma^2}$ of some 
given bandwidth $\sigma$, which works as a soft threshold for detecting high-contrast 
intensity edges in the image.
Thus, loss \eqref{eq:CRF_loss} encourages both the alignment of the segmentation boundary
to contrast edges in the (high-resolution) input image 
and the shortness/regularity of this boundary.

Weakly supervised segmentation methods may also use partial pixel-level ground truth where only 
some subset $Seeds\subset\Omega$ of image pixels has class labels \cite{kolesnikov2016seed,lin2016scribblesup,tang2018regularized}. In this case it is common to 
use {\em partial cross-entropy} loss
\begin{equation} \label{eq:pCE_loss}
   L_{pce}(S) \;\;=\;\; - \sum_{p\in Seeds} \ln S_p^{y_p}
\end{equation}
where $y_p$ is the ground truth label at a seed pixel $p$.

\subsection{Related balancing losses}
\label{sec: related balancing losses}
Segmentation and classification methods often use some ``balancing'' losses.
In the context of classification, image-level predictions can be balanced over the whole training data. For segmentation problems,
 pixel-level predictions can be balanced within each training image. Our loss is an example of size balancing. 
 Below we review some examples of related balancing loss functions used in prior work.   

\textbf{Fully supervised classification.} It is common to modify the standard cross-entropy loss 
to account for unbalanced training data where some classes are represented more than others. 
One common example is {\em weighted cross-entropy}, e.g. defined in \cite{belongie:cvpr2019} for \underline{image-level} predictions $S_i$ as 
\begin{equation} \label{eq:weightedce} 
    L_{wce}(S) \;\;  =\;\;-\sum_{i\in D} w_{y_i} \ln S_i^{y_i} 
\end{equation}
where class weights $w_k \propto \frac{1}{1-\beta^{v_k}} $ are motivated as a re-balancing factor based on the 
class distribution $v$ in the training dataset $D$ and $\beta$ is a hyper-parameter.
In the fully supervised setting, the purpose of re-weighting cross-entropy is not to make the predictions
even closer to the known labels, but to decrease over-fitting to over-represented classes, which improves the model's generality.

\textbf{Unsupervised classification.} In the context of clustering with soft-max models \cite{bridle1991unsupervised,krause2010discriminative}
it is common to use {\em fairness} loss encouraging equal-size clusters. 
In this case, there is no ground truth and fairness is one of the discriminative properties 
enforced by the total loss in order to improve the model predictions on unlabeled training data. 
The fairness was motivated by information-theoretic arguments in \cite{bridle1991unsupervised} deriving 
it as a negative entropy of the data-set-level {\em average prediction} 
$\hat{S}:=\frac{1}{|D|}\sum_{i\in D}  S_i$ for dataset $D$
\begin{align}
\label{eq:fairness_loss}
   L_{fair}(\hat{S}) \;\; & =\;\;  -H(\hat{S}) \;\;\equiv\;\; \sum_{k} \hat{S}^k \ln\hat{S}^k  \nonumber \\
   & \utc\;\;\;   \sum_{k} \hat{S}^k \ln\frac{\hat{S}^k}{1/K} \;\equiv\; KL(\hat{S}\|u) 
\end{align}
where $u = (\frac{1}{K},\dots,\frac{1}{K})$ is a uniform categorical distribution, and symbol $\,\utc\,$ indicates that 
the equality is up to some additive constant independent of $\hat{S}$.
Perona et al. \cite{krause2010discriminative} pointed out the equivalent KL-divergence formulation of the fairness in \eqref{eq:fairness_loss} and generalized it to a balanced partitioning constraint 
\begin{equation} \label{eq:balance_loss} 
    L_{bal}(\hat{S}) \;\;  =\;\;KL(\hat{S}\|v) 
\end{equation}
with respect to any given prior distribution $v$ that could be different from uniform. 

\textbf{Semantic segmentation with image-level supervision.}
Most weakly-supervised semantic segmentation methods use losses based on segment sizes. This is particularly true for
image-level supervision techniques \cite{kolesnikov2016seed,ismail:midl2018,araslanov2020single,pathak2015constrained}. Clearly, segments for tag classes 
should have positive sizes and segments for non-tag classes should have zero sizes.

Similarly to our paper, size-based constraints are typically defined for the image-level 
{\em average prediction} $\bar{S}$ computed from pixel-level predictions $S_p$, e.g. see \eqref{eq:ap}. 
Many generalized forms of pixel-prediction averaging can be found in the literature, 
where they are often referred to as {\em prediction pooling}. 
Some decay parameter often provides a wide spectrum of options from basic averaging 
to max-pooling. While the specific form of pooling matters, for simplicity, we discuss the corresponding balancing loss functions assuming basic average prediction $\bar{S}$, as in \eqref{eq:ap}.   

One of the earliest works on tag-supervised segmentation \cite{kolesnikov2016seed} 
uses {\em log-barriers} to ``expand'' tag objects in each training image
and to ``suppress'' the non-tag objects. Assuming image tags $T$, their {\em suppression loss} is defined as
\begin{equation} \label{eq:kolesnikov_suppress_loss} 
    L_{suppress}(\bar{S}) \;\;  \propto \;\;- \sum_{k\not\in T} \ln (1-\bar{S}^k) 
\end{equation}
encouraging each non-tag class to have zero average prediction $\bar{S}^k$, which implies zero predictions $S_p^k$ at each pixel.
Their {\em expansion loss} 
\begin{equation} \label{eq:kolesnikov_expand_loss} 
    L_{expand}(\bar{S}) \;\;  \propto \;\;- \sum_{k\in T} \ln \bar{S}^k. 
\end{equation}
encourages positive average predictions $S^k$ and non-zero volume for tag class segments.

We observe that the expansion loss \eqref{eq:kolesnikov_expand_loss} may have a bias to equal-size segments, which is
particularly evident in the case of average predictions. Indeed, as follows from \eqref{eq:kolesnikov_expand_loss} 
\begin{equation} \label{eq:bias}
    L_{expand}(\bar{S}) \;\;  \propto \;\; KL(\uT\|\bar{S}) 
\end{equation}
which is a special case of our size loss \eqref{eq:size_loss} when the size target $v=\uT$ is a uniform distribution over tag classes. 
The intention of the log barrier loss \eqref{eq:kolesnikov_expand_loss} is to push image-level size prediction $\bar{S}$ from the
boundaries of the probability simplex $\Delta_K$ corresponding to the zero-level for the tag classes $T$. 
Figure~\ref{fig:teaser}(a) shows the results for training based on 
the total loss combining CRF loss \eqref{eq:CRF_loss} with 
the log-barrier loss \eqref{eq:kolesnikov_expand_loss}. 
Its unintended bias to equal-size segments \eqref{eq:bias} is obvious. Note that
the mentioned decay parameter used for generalized average predictions should reduce such bias.

Alternatively, it may be safer to use barriers for $\bar{S}$ like 
\begin{equation} \label{eq:flat_bottom_barrier}
    L_{flat} \;\;=\;\;-\sum_{k\in T} \ln\max\{\bar{S}^k,\epsilon\}
\end{equation}
that have flat bottoms to avoid unintended bias to some specific size target inside the probability simplex $\Delta_K$. 
Similar thresholded barriers are commonly used, e.g. \cite{ismail:midl2018}.

\subsection{Contributions}

In general, it would be great to have effective image-level supervision for segmentation that only uses barriers
like \eqref{eq:kolesnikov_expand_loss} or \eqref{eq:flat_bottom_barrier} since they do not require any specific
size targets. This corresponds to tag-only supervision. 
However, our empirical results for semantic segmentation using such barriers were poor and comparable with those 
in \cite{kolesnikov2016seed}. A number of more recent semantic segmentation methods for tag-level supervision
have considerably improved such results \cite{ahn2018learning,zhou2022regional}, but they introduce significantly more complex multi-stage training procedures and various architectural modifications, which makes such methods hard to replicate, generalize, or to understand the results. 

We are focused on general easy-to-understand end-to-end training methods based on volumetric or size targets, which we show can be approximate. Our main point is that this image-level supervision requires little extra information on top of image tags, compared to full pixel-accurate ground truth, but it produces a training quality comparable with full supervision.
Approximate size targets is relatively easy to get from human annotators. Our work also motivates the development of
size predictor models that we plan to investigate in future work.

%% file: sec/2_properties.tex
\section{Size-target loss and its properties}
\label{sec:properties}
In contrast to our loss for image-level supervision of segmentation \eqref{eq:size_loss}, fairness \eqref{eq:fairness_loss} and balancing \eqref{eq:balance_loss} losses in discriminative clustering 
\cite{bridle1991unsupervised,krause2010discriminative,jabi2019deep} use the {\em reverse} KL divergence where the target distributions $u$ or $v$ appear as the second argument. As discussed later in the paper, this difference is important and loss \eqref{eq:balance_loss} is numerically unstable for image segmentation. 

Our proposed total loss is very simple
\begin{equation}
\label{eq: our loss1}
L_{total}\ :=\ L_{size} + L_{crf} 
\end{equation}
where the two terms come from our size target loss \eqref{eq:size_loss}  using the forward KL-divergence 
and standard CRF loss as in \eqref{eq:CRF_loss}.
The core new component is our size loss. 
Several important properties of $L_{size}$ are discussed below.

First of all, our size target loss \eqref{eq:size_loss} encourages specific approximate target volumes for
image tag classes, but also encourages zero volumes for not-tag classes. The CRF loss also contributes to the suppression of redundant classes. Unlike most prior work on image-level supervision for semantic segmentation,
e.g. \cite{kolesnikov2016seed,araslanov2020single,zhou2022regional}, we do not need separate suppression loss terms like 
\eqref{eq:kolesnikov_suppress_loss}. We validated this claim experimentally, 
they do not change the results.
\begin{figure}[!t]
    \centering
    \includegraphics[width=\linewidth]{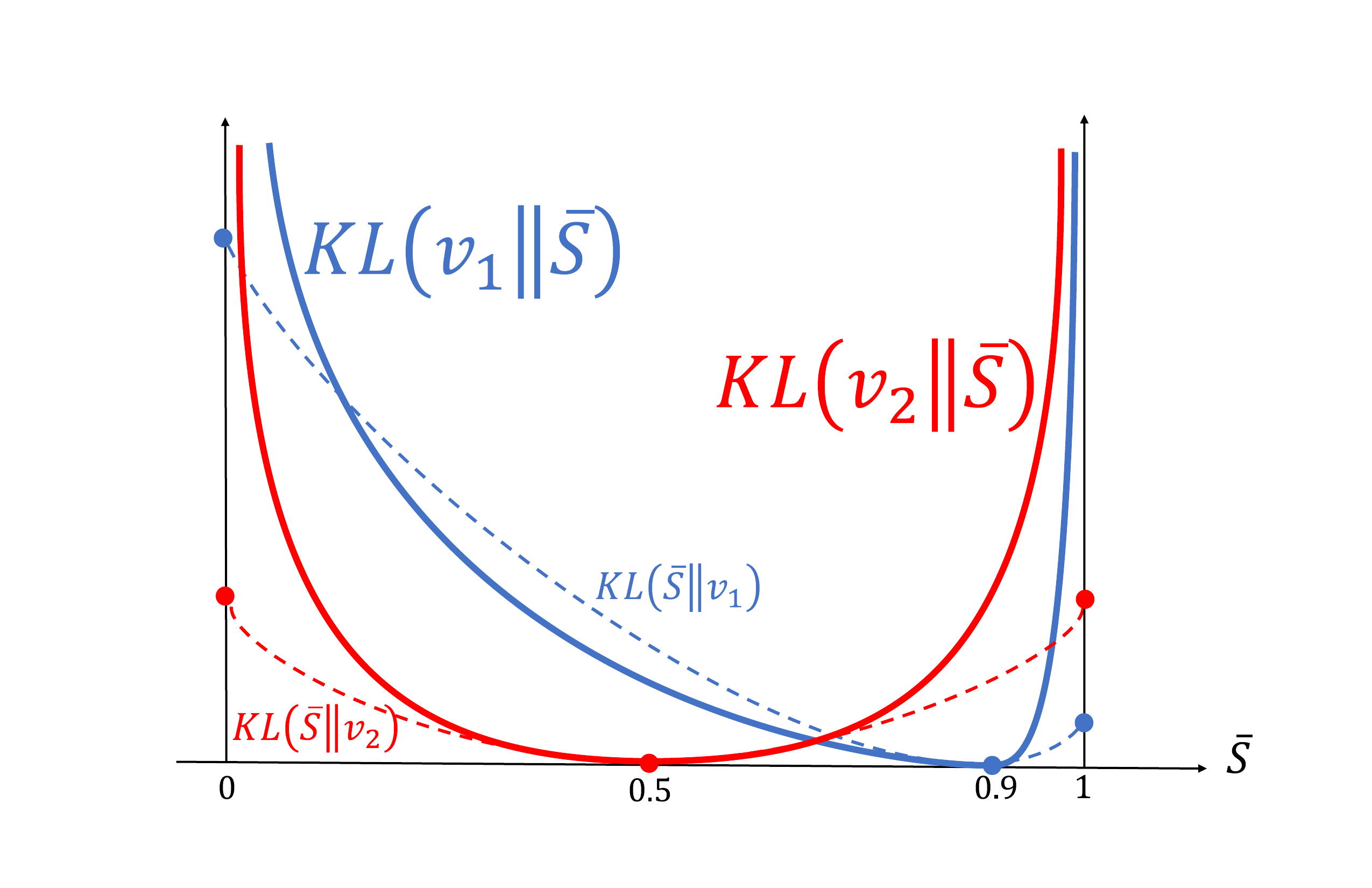}
    \caption{{\em Forward} vs {\em reverse} KL divergence. Assuming binary classification $K=2$, we can represent all possible probability distributions as points on the interval [0,1]. The solid curves illustrate our ``strong'' size constraint, i.e. the {\em forward} KL-divergence $KL(v\|\bar{S})$ for the average prediction $\bar{S}$. 
We show two examples of volumetric prior $v_1=(0.9,0.1)$ (blue curve) and $v_2=(0.5,0.5)$ (red curve). 
For comparison, the dashed curves represent reverse KL divergence $KL(\bar{S}\|v)$ 
commonly used in the prior art.}
    \label{fig: forward_reverse_kl}
\end{figure}

The size loss can also be integrated into other weakly-supervised settings, e.g. based on partial pixel supervision (seeds). Our size target loss \eqref{eq:size_loss} just needs to be combined with other losses used in these methods, e.g. \eqref{eq:pCE_loss}. In particular, approximate size targets significantly improve the results in \cite{tang2018regularized} in cases when seed scribbles are minimal, e.g. see the single click case in Fig.\ref{fig:scribble}.
\begin{equation}
\label{eq: our loss2}
L^{'}_{total}\ :=\ L_{size} + L_{crf} + L_{pce}
\end{equation}


As well-known, KL divergence is not symmetric with respect to its arguments. In the context of our work on image-level supervision
for semantic segmentation, the order of the estimate and target distributions matter significantly.
When the estimated distribution, 
e.g. $\bar{S}$, is the first argument, this loss is the reverse KL divergence as used in many prior arts on image classification. For classification problems, there is no numerical issue with putting the target distribution in the second position, because the target distribution usually does not have zero support for any class to enforce balance among classes, see \eqref{eq:balance_loss}. However, the image tags often only contain a small subset of the whole categories. This makes the reverse KL divergence ill-defined for the image-level target distribution since there will be many zeros at the denominator inside the logarithm. This motivates us to use the forward KL divergence which is more numerically stable. 

Moreover, the forward KL divergence has the so-called zero-avoiding property which is illustrated in Fig.\ref{fig: forward_reverse_kl}. The penalty will be infinite if any class corresponding to the non-zero target distribution gets zero prediction while the reverse KL divergence only enforces a finite penalty. Such property guarantees that we do not generate trivial solutions or miss any class labels from the image tag sets.



%% file: sec/3_experiments.tex
\section{Experiments}
\label{sec:experiments}
We conduct a comprehensive study to verify the effectiveness of our (approximate) size-target supervision and loss function. In Sec.\ref{sec:experiment_settings}, we detail our experimental settings. Sec.\ref{sec:image_level_exp} shows the results of image-level supervised semantic segmentation using our size-target approach. Sec.\ref{sec:scribble_exp} shows that our method can be seamlessly integrated into other weakly-supervised settings and boost the performance. Sec.\ref{sec:human} reports experiments conducted on human-annotated size targets. Sec.\ref{sec:compare_SOTA} tests our method on different network architectures and datasets, and compares our results with other state-of-the-art methods. Sec.\ref{sec: medical data} reports experimental results with medical images and compares our method with an alternative size constraint discussed in Sec.\ref{sec: related balancing losses}, called size barrier.   


\subsection{Experimental settings}
\label{sec:experiment_settings}
\textbf{Datasets.} We evaluate our approach on PASCAL VOC 2012 \cite{everingham2009pascal}, MS COCO 2014 segmentation dataset \cite{lin2014microsoft}, and 2017 ACDC Challenge\footnote[1]{https://www.creatis.insa-lyon.fr/Challenge/acdc/} medical dataset. The PASCAL dataset contains 21 classes. We adopt the augmented training set with 10,582 images \cite{hariharan2011semantic}, following the common practice \cite{chen2014semantic, kolesnikov2016seed}. Validation and testing contain 1449 and 1456 images. Scribble supervision of PASCAL dataset is from \cite{lin2016scribblesup}. MS COCO has 81 classes with 80K training and 40K validation images. The task of ACDC Challenge is to segment the left ventricular endocardium. We use the same experimental settings as \cite{ismail:midl2018}, where the training and validation sets contain 1674 and 228 images. Note that the exact size targets $\hat{v}= (\hat{v}_k)_{k=1}^K$ for all training images are available from the ground truth segmentation masks in all experiments. 

\textbf{Approximate size targets.} We train segmentation models using approximate size targets $v=(v_k)_{k=1}^K$ generated for each image either by human annotators or by corrupting the exact size targets $\hat{v}= (\hat{v}_k)_{k=1}^K$ with different levels of noise. In all cases, we report the obtained segmentation accuracy on validation data together with {\em mean relative error} (mRE) of the corresponding noisy size targets for the training data. For each training image containing class $k$, the {\em relative error} for the size target $v_k$ is defined as 
\begin{equation} \label{eq:RE}
RE(v_k) =  \frac{|v_k-\hat{v}_k|} {\hat{v}_k}
\end{equation}
where $\hat{v}_k$ is the exact (ground truth) size. mRE averages RE over all images and all classes, similarly to mIOU.
In case when approximate targets $v=(v_k)_{k=1}^K$ are set by human annotators, the corresponding relative size errors are computed directly from the definition \eqref{eq:RE}. See more in Sec.\ref{sec:human}.

When used, synthetic approximate targets $v=(v_k)_{k=1}^K$ are generated by corrupting the exact targets 
$\hat{v}= (\hat{v}_k)_{k=1}^K$
\begin{equation} \label{eq:synthetic_targets}
v_k \longleftarrow  (1+\epsilon) \hat{v}_k  \quad\text{for}\quad \epsilon\sim {\cal N}(0,\sigma)
\end{equation}
where $\epsilon$ is white noise with standard deviation $\sigma$ controlling the level of corruption and 
operator $\longleftarrow$ represents re-normalization making sure that corrupted targets $(v_k)_{k=1}^K$ still add up to one. 
Equation \eqref{eq:synthetic_targets} defines random variable $v_k$ as a function of $\epsilon$. Thus, in this case, mRE can be analytically estimated from $\sigma$ using \eqref{eq:RE} and \eqref{eq:synthetic_targets} as follows
\begin{equation} \label{eq:synthetic_RE}
mRE \;=\; E \left( \frac{|v_k-\hat{v}_k|}{\hat{v}_k} \right) \;\approx\; E(|\epsilon|) \;=\; \sqrt{\frac{2}{\pi}}\;\sigma
\end{equation}
where $E$ is the expectation operator. The approximation in the middle uses \eqref{eq:synthetic_targets} as an equality ignoring re-normalization of the corrupted sizes, and the last equality is a closed-form expression for 
the {\em mean absolute deviation} (MAD) of the Normal distribution ${\cal N}(0,\sigma)$.

\textbf{Evaluation metrics for segmentation.} We employ mean Intersection-over-Union (mIoU) as our evaluation criteria for PASCAL and COCO datasets. For the PASCAL test set, we use the official PASCAL VOC online evaluation server to assess our method. On the 2017 ACDC Challenge dataset, we adopt the mean Dice similarity coefficient (DSC) to measure the accuracy of our model.  

\textbf{Implementation details.} We evaluate our approach on PASCAL with two types of neural networks by concatenating the backbone of ResNet101 or the backbone of WR38 \cite{wu2016wider} with the decoder of DeepLabv3+ \cite{chen2018encoder}. For brevity, we call them ResNet101-based DeepLabv3+ and WR38-based DeepLabv3+ models. For experiments on COCO and ACDC datasets, we use ResNet101-based and MobileNetV2-based \cite{sandler2018mobilenetv2} DeepLabv3+, respectively. All backbone networks (ResNet-101, WR38, and MobileNetV2) are Imagenet \cite{deng2009imagenet} pretrained. The encoders (ResNet101, WR38, and MobileNetV2) and decoder (DeepLabv3+) used in our experiments are broadly used for semantic segmentation. 
PASCAL and COCO images are resized to 512 $\times$ 512 and medical images in ACDC are resized to 256 $\times$ 256 for training and testing. We augment the data with color jittering and horizontal flipping. Segmentation models are trained with stochastic gradient descent (SGD) for 60 epochs on PASCAL and COCO, and 200 epochs on ACDC. The learning rate scheduler is polynomial with a power of 0.9. The batch size is set to 16 for the Pascal dataset and 12 for the COCO and ACDC datasets. Initial learning rates are 0.005 for PASCAL and ACDC datasets. A lower learning rate of 0.0005 is used for the COCO dataset. 
we use loss function \eqref{eq: our loss1} for image-level-supervised semantic segmentation and \eqref{eq: our loss2} for pixel-level supervision. The implementation of CRF loss \eqref{eq:CRF_loss} is the same as \cite{tang2018regularized}. Size-target loss and pCE are used for medical images, see Sec.\ref{sec: medical data} for details. 


\subsection{Image-level-supervised segmentation}
\label{sec:image_level_exp}
\begin{figure}[!t]
    \centering
    {\includegraphics[width=\linewidth]{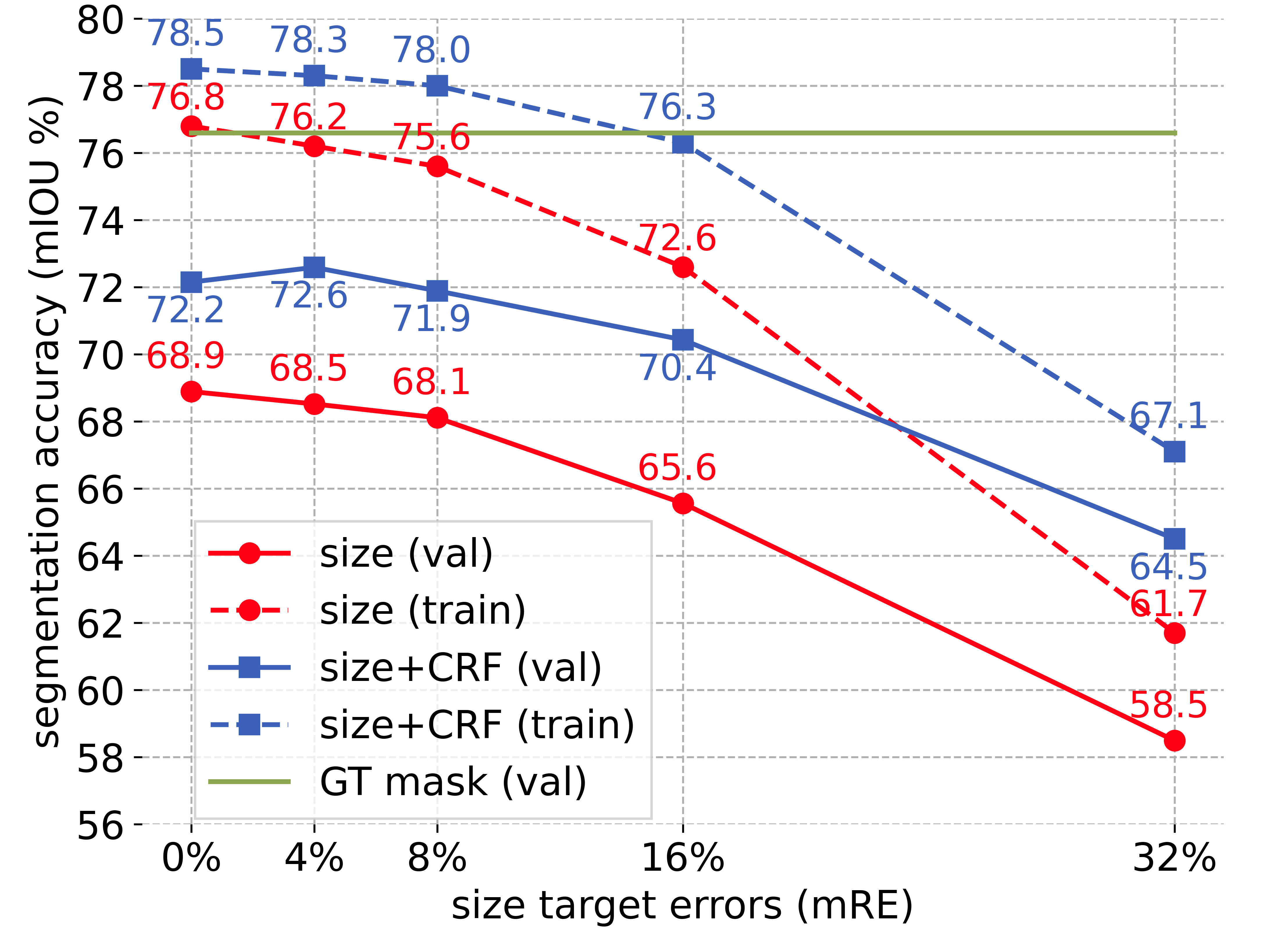}} 
    \caption{Segmentation accuracy for our approximate size-target supervision on PASCAL's training and validation data. 
    The segmentation is trained using losses \eqref{eq:size_loss} (red) or \eqref{eq: our loss1} (blue), 
    where size targets are subject to various levels of corruption (\ref{eq:synthetic_targets},\ref{eq:synthetic_RE}).} 
    \label{fig:tag}
\end{figure}
Precise size-targets might be as difficult to obtain as precise masks. However, our approach works very well with approximate size targets, which are much easier to get. We evaluate robustness to size errors as follows. We train the ResNet101-based DeepLabv3+ segmentation model with synthetic size targets \eqref{eq:synthetic_targets} using five levels of corruption ($mRE \in \{0\%,4\%,8\%,16\%,32\%\}$), where $mRE=0\%$ means exact size targets. The conversion between mRE and $\sigma$ is described in \eqref{eq:synthetic_RE}. The results in Figure~\ref{fig:tag} demonstrate the robustness of our approach to size errors. The mIOU scores on the PASCAL validation set (red solid line) only drop slightly when mRE \eqref{eq:synthetic_RE} is within 16\%. With the help of CRF loss \eqref{eq:CRF_loss}, the segmentation model (solid blue line) is more resilient to noise. When $mRE=4\%$, there is a noticeable increase in accuracy. As the training accuracy curve is monotonically decreasing (dashed blue line), the increase in accuracy on the validation set may be attributed to a better generalization of the neural networks. 


\subsection{Scribble-supervised segmentation}
\label{sec:scribble_exp}
\begin{figure}[!t]
    \centering
    {\includegraphics[width=\linewidth]{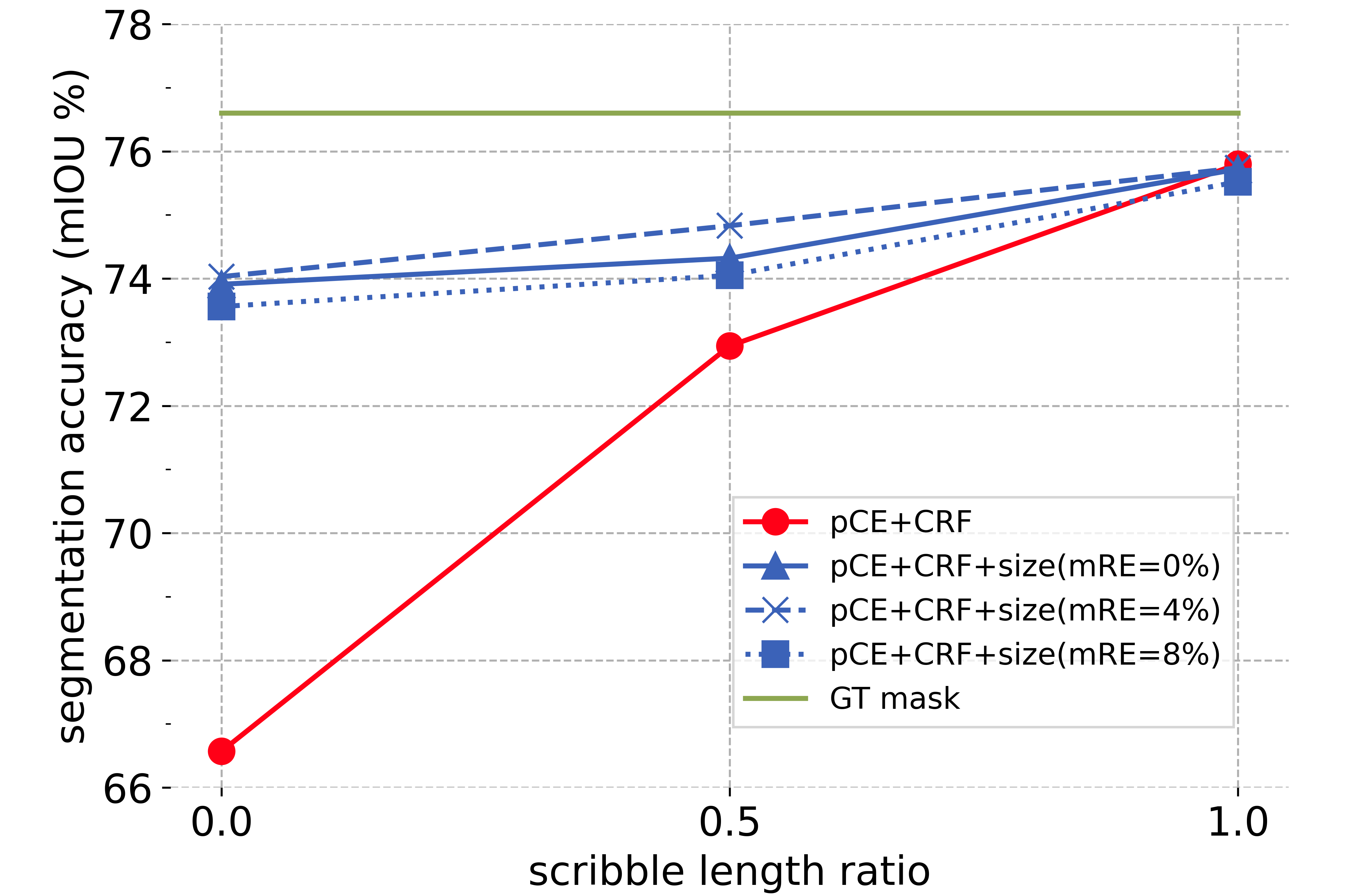}} 
    \caption{Segmentation accuracy for scribble supervision {\color{blue} with} and {\color{red} without} 
    our approximate size targets on PASCAL's validation data.}
    \label{fig:scribble} \vspace{-2ex}
\end{figure}

Our size-target approach can be seamlessly integrated with partial pixel supervision (seeds/scribbles) for semantic segmentation. 
Figure~\ref{fig:scribble} reports the quantitative results of scribble-supervised semantic segmentation with or without our method using ResNet101-based DeeplabV3+. The results show that our method can significantly enhance the training, especially when the supervision is weak. As shown in Fig.\ref{fig:scribble}, the performance of the segmentation model trained without our method degrades dramatically with the reduction of scribble length (red line). 
Note that $\text{scribble length ratio} = 0.0$ means only one pixel is labeled for each object in the image (one click). Using approximate size targets and minimal localization cues (one click), the segmentation model's performance closely approaches that achieved with full scribble supervision.


\subsection{Human-annotated size targets}
\label{sec:human}
\begin{figure}[!t]
    \centering
    {\includegraphics[width=\linewidth]{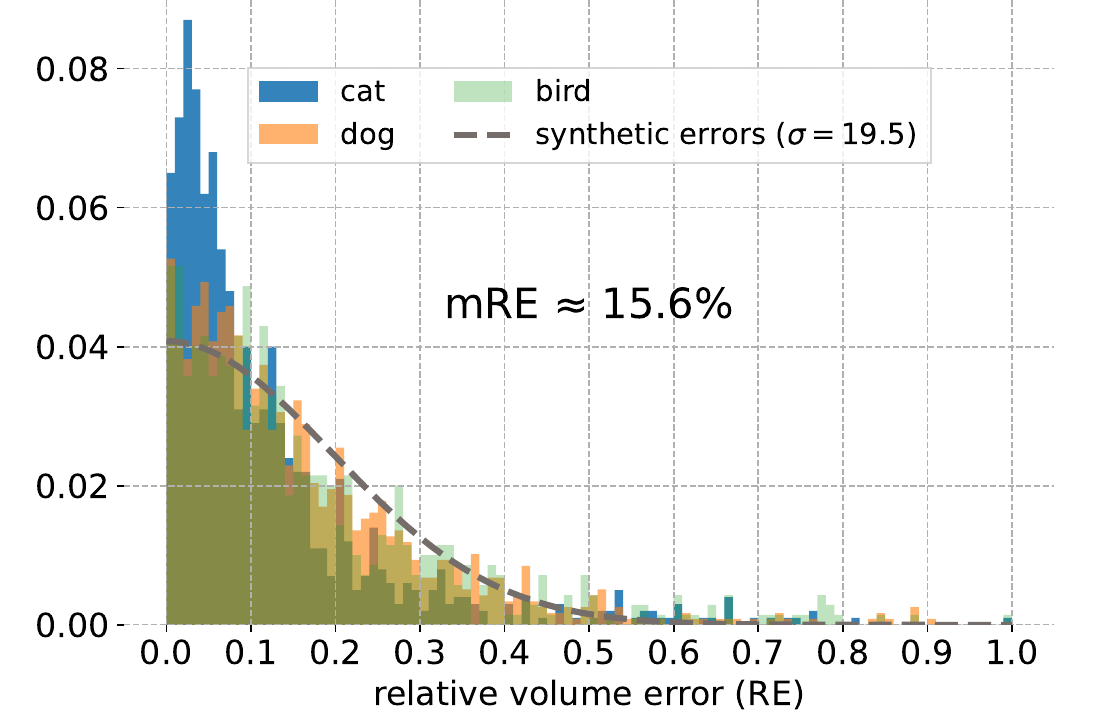}} 
    \caption{Human annotation quality: relative errors histograms on PASCAL VOC classes (dog, cat, and bird). The histograms are normalized over the image count in each class. The relative size errors \eqref{eq:RE} average to $mRE= 15.6\%$. For comparison, the dashed line shows the distribution of relative errors for our synthetic corruption of size targets \eqref{eq:synthetic_targets} for $\sigma=19.5$ corresponding to the same $mRE= 15.6\%$ \eqref{eq:synthetic_RE}.}
    \label{fig:histogram} \vspace{-2ex}
\end{figure}

\textbf{Annotation tool.} Experiments in Sec.\ref{sec:image_level_exp} and Sec.\ref{sec:scribble_exp} are conducted with synthetic size targets. To validate the usefulness of our approach in real-world settings, we annotated training images in the subset of PASCAL classes (cat, dog, bird). We implemented an annotation interface including an assistance tool. The tool overlays grid lines partitioning the image into $5\times4$ small rectangles or $3\times3$ large rectangles. The user can enter the size for a given class in each image either by counting rectangles (fractions are OK) or by directly entering the corresponding percent with respect to the image size. Annotators can interactively select the finer or coarser partitioning for each image depending on the object size. We evaluate relative errors \eqref{eq:RE} for human annotations. 
Empirical evidence shows that annotators are approximately two times more accurate when they employ the assistance tool, especially for small objects in the image.  The last two columns of Table \ref{tab:human} report the annotation speed\footnote[2]{According to \cite{Andriluka_2018}, it takes on average 19 minutes 
to fully annotate an image in the COCO dataset and 1.5 hours for one Cityscapes image \cite{Cordts2016Cityscapes}.} per image and mean relative error \eqref{eq:RE} for each individual class. Figure \ref{fig:histogram} shows the histogram of the relative errors made by annotators for each class, compared with the distribution of relative errors for our synthetic corruption of size targets. Note that the mean relative error (15.6\%) reported in Fig.\ref{fig:histogram} is the weighted average of the mean relative errors for the three classes listed in Table \ref{tab:human}. 

\begin{figure}[!t]
    \centering
    {\includegraphics[width=\linewidth]{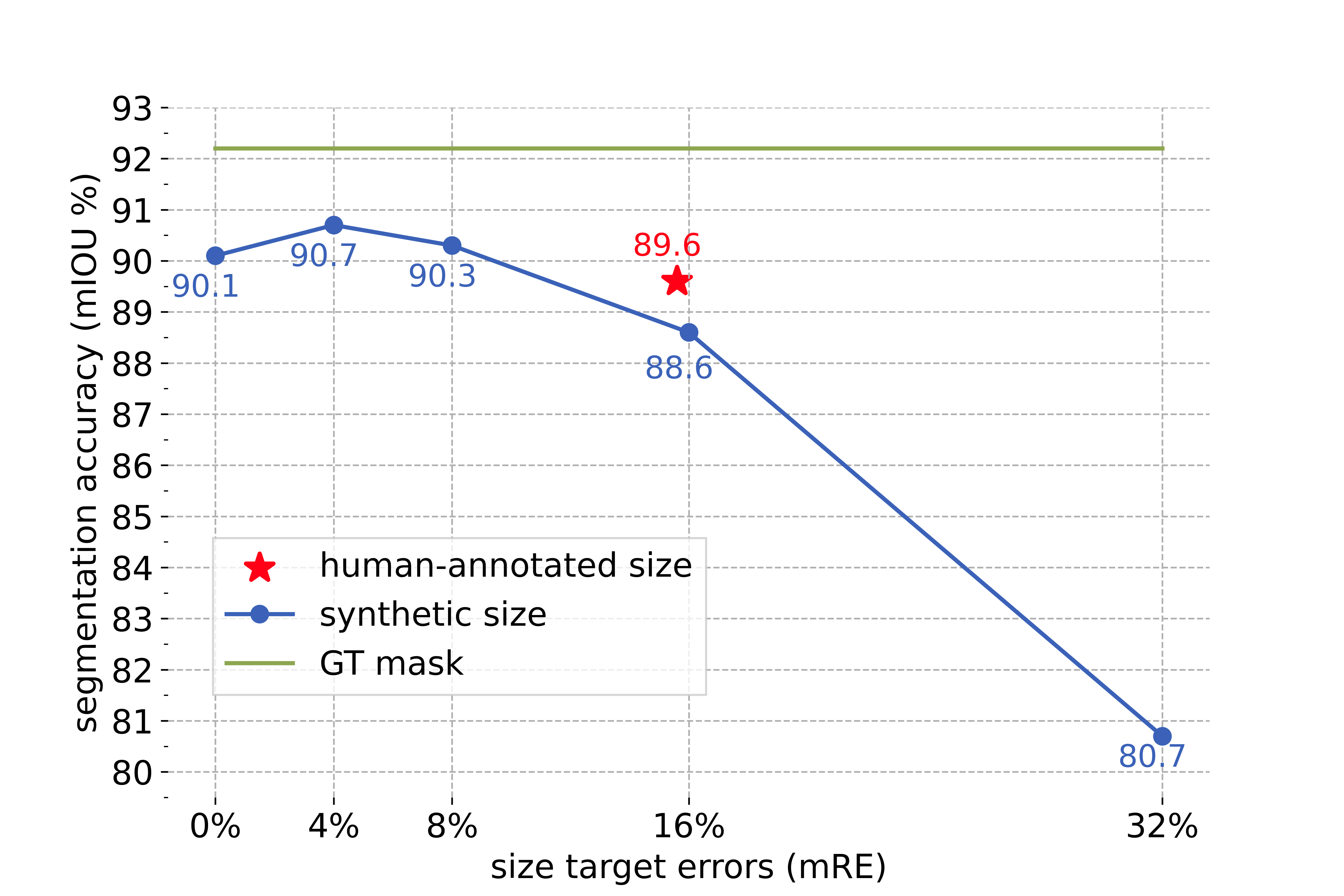}} 
    \caption{Segmentation accuracy for size-target supervision with human-defined and 
    synthetically corrupted size targets on a subset of PASCAL classes (cat, dog, bird). 
    Unlike Table \ref{tab:human}, we train multi-class segmentation here.
    The red star corresponds to human-annotated sizes ($mRE = 15.6\%$) and the blue curve to 
    synthetic size targets with different levels of corruption (\ref{eq:synthetic_targets},\ref{eq:synthetic_RE}).
    The green line is the segmentation accuracy for full GT mask supervision. }
    \label{fig:human}  \vspace{-2ex}
\end{figure}

\textbf{Training with human-annotated size.} 
We use the human-annotated size targets to train multi-class semantic segmentation models (ResNet101-based DeeplabV3+) over all images of three classes (cat, dog, bird) in PASCAL dataset. For comparison, Figure~\ref{fig:human} presents the accuracy of the 4-way (background, cat, dog, and bird) semantic segmentation models trained with human-annotated size, five noise levels of synthetic size targets \eqref{eq:synthetic_targets}, and ground truth masks. 
\begin{table}[!t]
\begin{center}
\begin{tabular}{c|cc|ccc}
\hline
\scriptsize{supervision} & gt mask & gt size & \multicolumn{3}{c}{human-annotated size} \\
\hline
            & mIOU    & mIOU    & mIOU   & speed & mRE \\
\hline
cat         & \cellcolor{gray!15}90.6\%  & \cellcolor{gray!15}88.8\%  & \cellcolor{gray!15}88.0\% & 12.6s & 12.3\% \\
dog         & \cellcolor{gray!15}88.1\%  & \cellcolor{gray!15}84.3\%  & \cellcolor{gray!15}84.5\% & 9.1s  & 16.6\% \\
bird        & \cellcolor{gray!15}88.8\%  & \cellcolor{gray!15}86.2\%  & \cellcolor{gray!15}86.4\% & 15.2s & 20.1\% \\
\hline
\end{tabular}
\end{center} \vspace{-2ex}
\caption{Human-annotated size targets for segmentation. The last two columns show 
the average speed and relative errors (mRE) we obtained for human annotation of size targets for
individual classes (cat, dog, bird) on PASCAL. The shaded cells compare the accuracy (mIOU) 
of binary segmentation (of each class separately) trained by human-labeled size, exact size, and ground truth mask.}
\label{tab:human} \vspace{-2ex}
\end{table}
Figure~\ref{fig:human} shows that the accuracy of the segmentation model trained with the human-annotated size is on par with the exact size-target supervision. As shown in Fig.\ref{fig:histogram}, human-produced size errors are mostly below $10\%$, but occasional large mistakes (heavy tails) raise the mean error. The red point in Fig.\ref{fig:human} shows the robustness of the segmentation model to such ``heavy tail'' size errors.  

To further evaluate our approach, we also train {\em binary} semantic segmentation models with images of each individual class (cat, dog, or bird) using the corresponding human-annotated size, exact size, and full ground truth masks. Accuracy on the validation set is presented in the shaded cells in Table \ref{tab:human}. 
Despite the occasional significant relative error in human annotations, the accuracy of models trained with such annotations is comparable to, or even exceeds, that achieved with precise size targets. This shows the potential of our approach for practical applications where image-level human annotation of approximate 
sizes could be significantly less expensive compared to pixel-level masks. 

\begin{table}[!b]
\begin{center}
\small
\begin{tabular}{c|cccccc}
\hline
\rowcolor{gray!15}
\multicolumn{7}{c}{PASCAL Dataset} \\
\hline
Methods & Backbone & Sup. & Val & Test & S & M \\
\hline
 RCA \cite{zhou2022regional}  & WR38 & $\mathcal{I}$, $\mathcal{S}$ & 72.2 & 72.8 & & \checkmark \\
 PPC \cite{du2022weakly}  & R101 & $\mathcal{I}$, $\mathcal{S}$ & 72.6 & 73.6 & & \checkmark \\
 URN \cite{li2022uncertainty}   & R101 & $\mathcal{I}$ & 69.5 & 69.7 & & \checkmark \\
 SANCE \cite{li2022towards} & R101 & $\mathcal{I}$ & 70.9 & 72.2 & & \checkmark\\
 \hline
 SSSS \cite{araslanov2020single}  & WR38 & $\mathcal{I}$ & 62.7 & 64.3 & \checkmark &\\
 AFA \cite{ru2022learning}   & MiT-B1 & $\mathcal{I}$ & 66.0 & 66.3 & \checkmark &\\
 \textbf{Ours}  & R101   & $\mathcal{V}$ 0\% & 72.2 & 70.8 & \checkmark &\\
 \textbf{Ours}   & WR38   & $\mathcal{V}$ 8\% & 72.7 & 71.6 & \checkmark &\\
\hline
\rowcolor{gray!15}
\multicolumn{7}{c}{COCO Dataset} \\
\hline
 URN \cite{li2022uncertainty} & R101 & $\mathcal{I}$ & 40.7 & & & \checkmark\\
 SANCE \cite{li2022towards} & R101 & $\mathcal{I}$ & 44.7 & & & \checkmark\\
 \hline
AFA \cite{ru2022learning} & MiT-B1 & $\mathcal{I}$ & 38.9 & & \checkmark &\\
\textbf{Ours}   & R101   & $\mathcal{V}$ 8\% & 45.0 & & \checkmark & \\
\hline
\end{tabular}  \vspace{-3ex}
\end{center}
\caption{Segmentation accuracy (mIOU\%) on PASCAL validation and test sets, and COCO validation set. $\mathcal{I}$, $\mathcal{S}$, and $\mathcal{V}$ denote image-level, saliency, and size (volume) target supervision, respectively. The percentage next to ``$\mathcal{V}$'' indicates the accuracy of size targets (mRE) used for corruption (\ref{eq:synthetic_targets},\ref{eq:synthetic_RE}). Columns ``S'' and ``M''  represent the single-stage and multi-stage training schemes.}
\label{tab:SOTA}
\end{table}

\subsection{Comparison with state-of-the-arts}
\label{sec:compare_SOTA}
\begin{figure}[!b]
\includegraphics[width=0.98\linewidth]{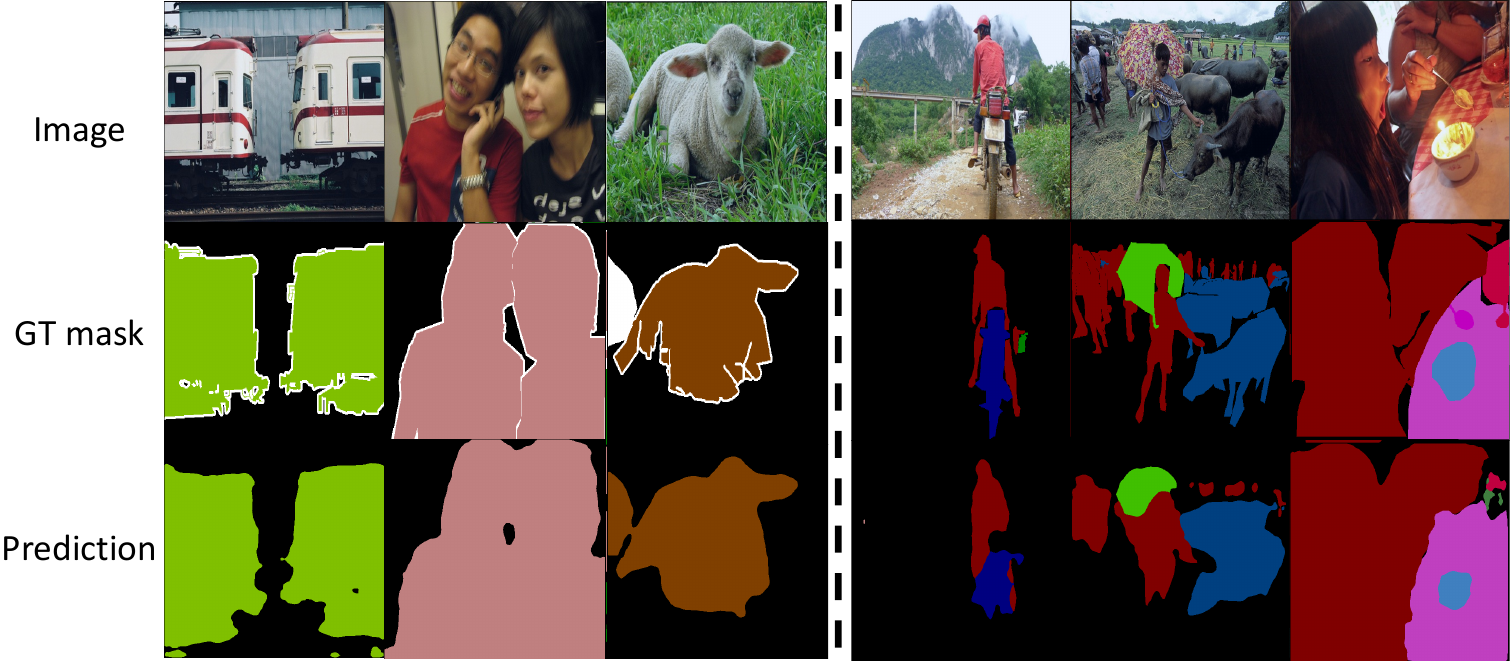} 
    \caption{Qualitative segmentation examples for supervision by size-targets ($mRE=8\%$) 
    on PASCAL with WR38-based model (left) and COCO with Resnet101-based model (right) .}
    \label{fig:qualitative} \vspace{-2ex}
\end{figure}

Our method is simple and generally applicable to any semantic segmentation networks of arbitrary architectures 
in an end-to-end manner. Besides ResNet101-based model used in Sec.\ref{sec:image_level_exp}-\ref{sec:human},  We also evaluate size-target supervision with WR38-based DeepLabv3+ on the PASCAL dataset. Our approach is also 
tested on the COCO 2014 data using ResNet101-based DeepLabv3+ model. Figure~\ref{fig:qualitative} presents the qualitative examples of our method on PASCAL (left) and COCO (right) validation sets.
The upper part of Table \ref{tab:SOTA} reports the accuracy of semantic segmentation models trained on the exact and synthetic size targets ($mRE=8\%$) with our loss \eqref{eq: our loss1} on PASCAL, compared with other state-of-the-art methods separated by training schemes. Our proposed method surpasses other single-stage methods by a large margin with a tiny amount of extra information. Our approach is comparable to, if not outperforming, much more intricate multi-stage methods. 
The lower part of Table \ref{tab:SOTA} shows the results of our method on the COCO dataset. Similar to the results on PASCAL, our approach significantly outperforms other single-stage methods and is even superior to complex multi-stage approaches. 


\subsection{Medical data: size-target vs. size-barrier}
\label{sec: medical data}
In Sec.\ref{sec:image_level_exp}-\ref{sec:human}, we demonstrate the effectiveness of our approach to natural images. We believe that our method offers even more potential in the domain of medical image segmentation, where the object volumes within similar types of medical images are consistent and easy for healthcare professionals to estimate. We evaluate our method with medical images on the ACDC dataset and compare our size-target method with thresholded size-barrier \cite{kolesnikov2016seed,ismail:midl2018} discussed in Sec.\ref{sec: related balancing losses}.
\begin{figure}[!t]
    \centering
    {\includegraphics[width=\linewidth]{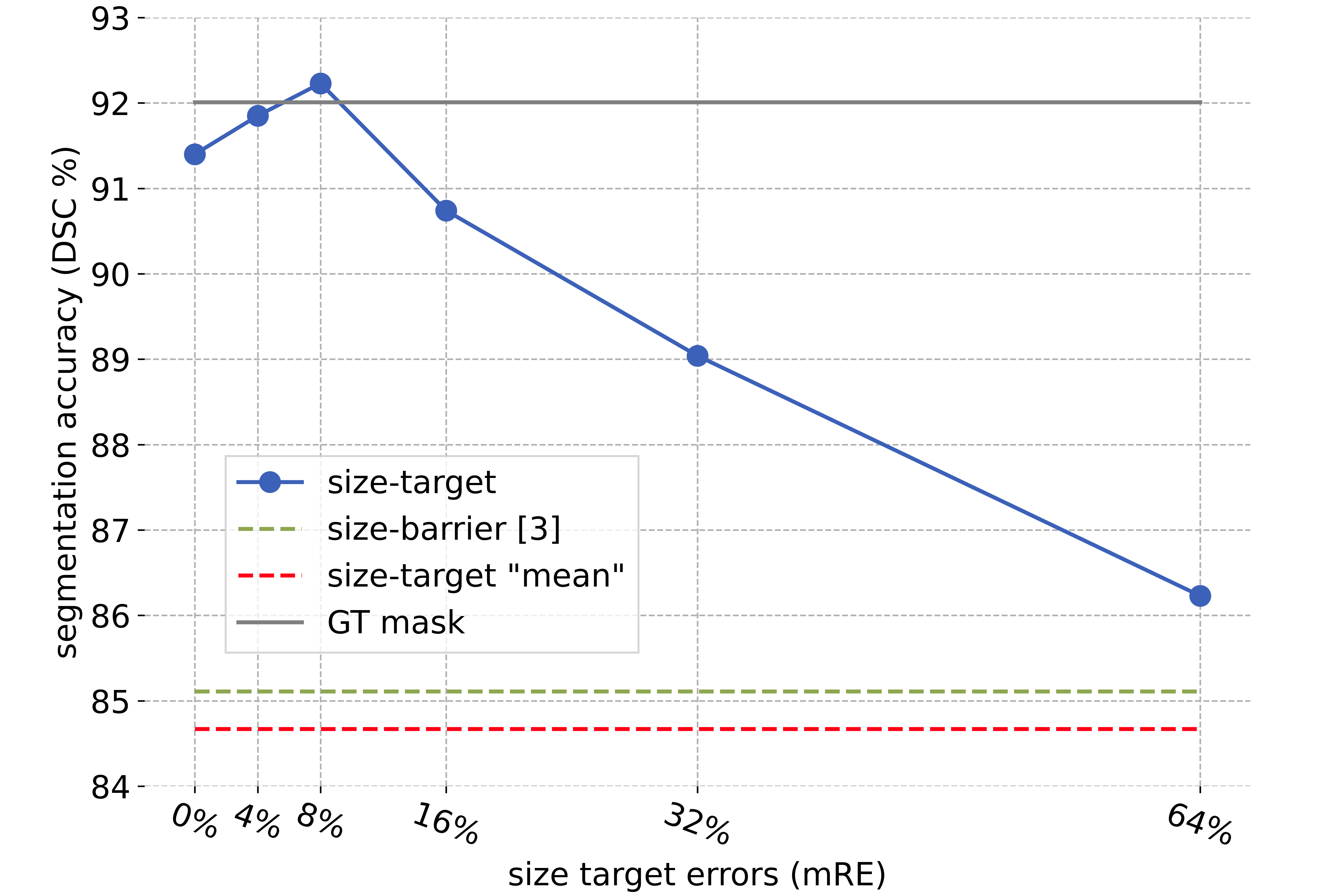}} 
    \caption{Size-targets \eqref{eq:size_loss} vs. size-barriers \eqref{eq:barrier_loss} on ACDC dataset. 
    The accuracy of the segmentation models (MobileNetV2-based DeeplabV3+) is measured by {\em Dice coefficient} (DSC). 
    The blue curve corresponds to size-targets of different accuracy. The green line corresponds to
    the quadratic size-barrier \eqref{eq:barrier_loss} in \cite{ismail:midl2018}.
    The red line shows the accuracy for the model trained by \eqref{eq:size_loss} using a fixed size-target for all training images (the average object size).  
    \label{fig:target_vs_barrier_medical}}  \vspace{-2ex}
\end{figure}

Size barriers are naturally used to suppress the predictions of non-tag classes and enforce positive sizes for tag classes. For example, \cite{kolesnikov2016seed} expand the size of tag classes using {\em log-barrier} \eqref{eq:kolesnikov_expand_loss}. 
However, as we have discussed in Sec.\ref{sec: related balancing losses}, such losses can lead to undesirable uniform volumetric bias, see \eqref{eq:bias}. Thresholded barriers, e.g. \eqref{eq:flat_bottom_barrier}, avoid this bias by enforcing some lower bound on the object size, assuming it is known. We compare our size-target loss \eqref{eq:size_loss} with the thresholded quadratic size-barrier loss \cite{ismail:midl2018}
\begin{equation} \label{eq:barrier_loss}
   L_{flat\_sq}(S) \;\;=\;\; \sum_k \left(\max\{a_k - \Bar{S}^k,0\}\right)^2
\end{equation}
where $a_k$ is a lower {\em size barrier} for class $k$ \cite{ismail:midl2018}.

Comparative experiments are conducted in the context of binary medical image segmentation. Segmentation models are trained with pCE loss \eqref{eq:pCE_loss} on seeds and size-target \eqref{eq: our loss1} or size-barrier \eqref{eq:barrier_loss}.
Seeds used in the experiments are obtained using the same method provided in \cite{ismail:midl2018}. The object and background size barrier, $a_{obj}$ and $a_{bg}$, used in the size-barrier loss \eqref{eq:barrier_loss} are set to the same values as in \cite{ismail:midl2018}. 
In the size-barrier experiments, similarly to \cite{ismail:midl2018}, we suppress the object class if it is absent in the image, using the loss function $ L_{sup}(S) = (\Bar{S}^{obj})^2$. Conversely, the size-target loss automatically suppresses absent classes, see Sec. \ref{sec:properties}.
\begin{figure}[!t]
    \centering
    {\includegraphics[width=0.8\linewidth]{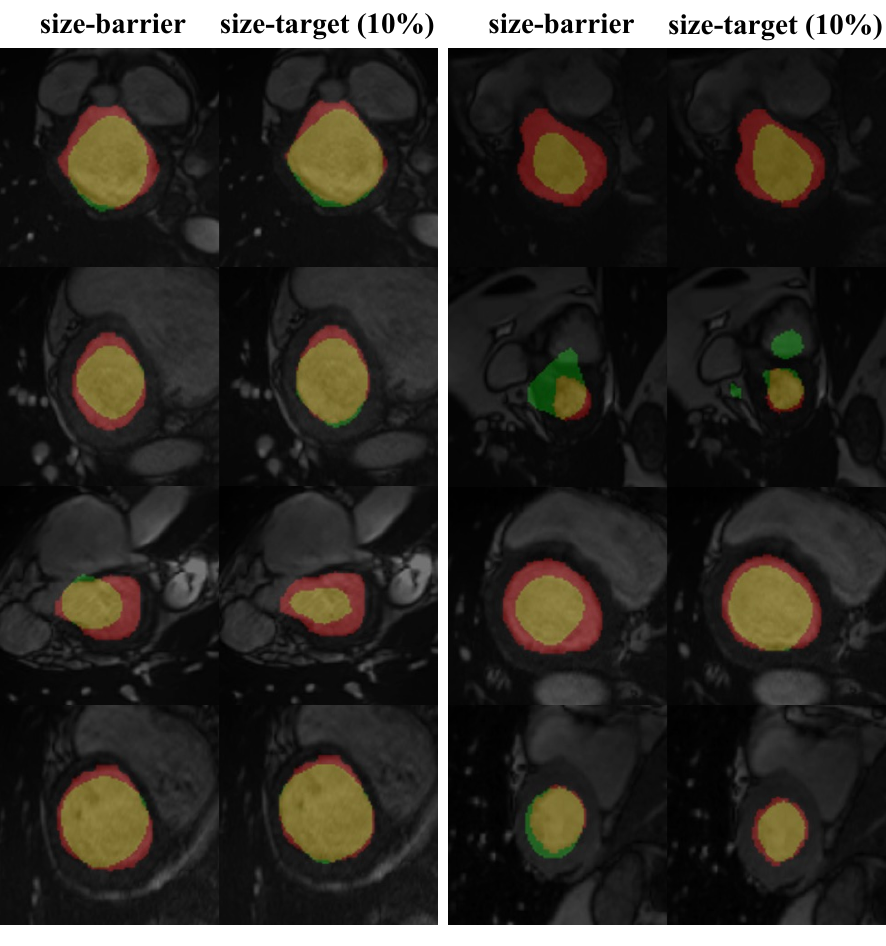}} 
    \caption{Randomly selected results comparing size-barrier \eqref{eq:barrier_loss} in \cite{ismail:midl2018} and 
    our approximate size targeting \eqref{eq:size_loss} with $mRE=8\%$. 
    Yellow shows true positive pixels, green is false positive, and red is false negatives. Size barrier results tend to under-segment the objects, which may explain the differences shown in Fig.\ref{fig:target_vs_barrier_medical}. 
    \label{fig:medical_examples}}  \vspace{-2ex}
\end{figure}

Qualitative predictions of (thresholded) size-barrier and size-target are presented in Fig.\ref{fig:medical_examples}. Figure~\ref{fig:target_vs_barrier_medical} displays the accuracy of binary semantic segmentation models trained using pCE, jointly with either thresholded size barriers or size targets. Similar to experiments in Sec.\ref{sec:image_level_exp} and Sec.\ref{sec:human}, we use size targets corrupted by \eqref{eq:synthetic_targets} ($mRE \in \{0\%,4\%,8\%,16\%,32\%,64\%\}$). As shown in Fig.\ref{fig:target_vs_barrier_medical}, size-target loss consistently outperforms size-barrier loss even when the mRE is 64\%. A peak is also observed in Fig.\ref{fig:target_vs_barrier_medical} in the accuracy curve, which is consistent with Fig.\ref{fig:tag} and Fig.\ref{fig:human}. Compared to Fig.\ref{fig:tag}, our size-target approach is even more robust to the size errors on medical images. Possibly, the lower variability of object sizes on medical data (vs. PASCAL) makes training simpler. The lower variability of object sizes allows us to train a segmentation model even with the fixed mean size target of the training images and achieve decent performance, see red line in Fig.\ref{fig:target_vs_barrier_medical}.

%% file: sec/4_conclusions.tex
\section{Conclusions}
\label{sec:conclusions}
We proposed a new image-level supervision for semantic segmentation: size targets. Such targets could be approximate and some level of errors benefit generalization.
The size annotation by humans requires little extra effort on top of the standard image-level class tags
and it is much cheaper than the full pixel-accurate ground truth masks. We proposed an effective size-target loss based on KL divergence between the soft size-targets and the average prediction. In combination
with the standard CRF-based regularization loss, our approximate size-target supervision achieves 
segmentation accuracy comparable with full pixel-precise supervision and outperforms most state-of-the-art
image-level supervision methods that are significantly more complex.
